\documentclass{article}

\usepackage[nonatbib,preprint]{neurips_2021}

\usepackage[utf8]{inputenc} 
\usepackage[T1]{fontenc}    
\usepackage{hyperref}       
\usepackage{url}            
\usepackage{booktabs}       
\usepackage{amsfonts}       
\usepackage{nicefrac}       
\usepackage{microtype}      
\usepackage{xcolor}         
\usepackage{graphicx}
\usepackage{multirow}
\usepackage{tabularx}
\usepackage{subfigure}
\usepackage{caption}
\newenvironment{corollary}[2][Corollary]{\begin{trivlist}
\item[\hskip \labelsep {\bfseries #1}\hskip \labelsep {\bfseries #2.}]}{\end{trivlist}}

\newcommand{\AAA}{AdvAA}

\title{Challenges of Adversarial Image Augmentations}
\author{%
    Arno Blaas \\
    University of Oxford\\
    \texttt{arno@robots.ox.ac.uk} \\
    \And
    Xavier Suau \\
    Apple\\
    \texttt{xsuaucuadros@apple.com} \\
    \And
    Jason Ramapuram \\
    Apple \\
    \texttt {jramapuram@apple.com} \\
    \And
    Nicholas Apostoloff \\
    Apple \\
    \texttt{napostoloff@apple.com}\\
    \And
        Luca Zappella \\
    Apple\\
    \texttt{lzappella@apple.com}
}

\begin{document}

\maketitle

\begin{abstract}
Image augmentations applied during training are crucial for the generalization performance of image classifiers. Therefore, a large body of research has focused on finding the optimal augmentation policy for a given task. Yet, RandAugment \cite{cubuk2020randaugment}, a simple random augmentation policy, has recently been shown to outperform existing sophisticated policies. Only Adversarial AutoAugment (\AAA) \cite{zhang2019adversarial}, an approach based on the idea of adversarial training, has shown to be better than RandAugment. In this paper, we show that random augmentations are still competitive compared to an optimal adversarial approach, as well as to simple curricula, and conjecture that the success of \AAA{} is due to the stochasticity of the policy controller network, which introduces a mild form of curriculum. 
\end{abstract}

\section{Introduction}
\label{sec:Introduction}
The success of a deep vision model can greatly depend on the data augmentations that were used during training. For example, when training a WideResNet-28-10 on CIFAR10 \cite{krizhevsky2009learning}, we observed a test set accuracy increment of $6.8\%$ when using augmentations. For natural images, these augmentations can be categorized as geometric (e.g, horizontal flips, translations), color (e.g., histogram equalization), and distortion (e.g., random cropping, or random greypainting of parts of the images, commonly referred to as Cutout \cite{devries2017improved}).

Finding the right augmentation policy, which determines the type and magnitude of the augmentation to use at a given training point, has been the subject of a large amount of research. Novel approaches  range from using reinforcement learning \cite{cubuk2019autoaugment}, Bayesian optimization \cite{lim2019fast}, or evolutionary algorithms \cite{ho2019population} to only name a few. While these methods managed to surpass the performance of existing hand-crafted augmentation baselines, they were subsequently outperformed by RandAugment~\cite{cubuk2020randaugment}, a simple random augmentation policy. The only exception is Adversarial AutoAugment~\cite{zhang2019adversarial} (\AAA), which uses an adversarial approach to guide the search for the right augmentation policy and significantly outperforms RandAugment. Yet, it is unclear to which extent the polices selected by \AAA{} are adversarial; hence, the question remains: do adversarial augmentations truly help?

In this work, we shed more light on this open question and show that exactly solving the adversarial objective in \AAA{} is detrimental to generalization. We also show that curricula that increase the adversarial degree of the augmentation is only as effective, if not less, than a random policy. The comparison between the policies selected by \AAA{} and those selected by an exhaustive search suggests that the success of \AAA{} might lie in the implicit randomness of the solution to the optimization of the adversarial objective, as we further explain in the Section \ref{sec:Discussion}. We conclude our work with a glimpse of hope suggesting a new direction that has interesting preliminary results and could be a founding principle for new policies.

\section{Background}
\label{sec:background}

AutoAugment \cite{cubuk2019autoaugment} is a natural starting point for an overview of recent works on automating the search for the ideal augmentation policy, as it spans the policy search space that many subsequent works adopt for their own policy search. The basis of AutoAugment's search space consists of $16$ operations (ShearX/Y, TranslateX/Y, Rotate, AutoContrast, Invert, Equalize, Solarize, Posterize, Contrast, Color, Brightness, Sharpness, Cutout \cite{devries2017improved}, Sample Pairing \cite{inoue2018data}). 
The search space for augmentation policies is then spanned by pairs of these operations, where each operation has a probability and magnitude along a uniform spacing of $11$ probabilities, within the range $[0, 1]$, and $10$ magnitudes, within the range $[0, \mathrm{maxmagnitude}]$.
At each training step, $5$ of such policies are sampled with equal probability leading to a search space of $(16*11*10)^{2 \times 5} \approx 10^{32}$ elements.
In AutoAugment, the authors aim to find the optimal set of augmentations by using a reinforcement learning approach, where the validation set accuracy is used as reward. The augmentation policies they find improve test set accuracy by $0.4\%$ and $1.3\%$ on CIFAR10 and CIFAR100 \cite{krizhevsky2009learning} when using a Wide-ResNet-28-10.
Lim et al. \cite{lim2019fast} drastically shorten the duration of the policy search by using a Bayesian optimization approach while maintaining a similar performance. 
Hataya et al. \cite{hataya2020faster} shorten the duration even further by transforming the search into an online search (that is, during training of the final model). The authors propose to minimize the Wasserstein distance between the augmented and the original training data empirical distribution, again while maintaining a comparable performance in terms of test set accuracy. 
Ho et al. \cite{ho2019population} also search for the optimal augmentation policy in an online fashion using an evolutionary algorithm (population-based training \cite{jaderberg2017population}), reducing the search time and even outperforming the original AutoAugment on CIFAR100 when using a Wide-ResNet-28-10.

However, all of these sophisticated approaches were outperformed by RandAugment, a simple random augmentation policy introduced in \cite{cubuk2020randaugment} which uniformly samples the components of operation pairs to be applied and uses either a constant or a random magnitude (to the same effect, as shown in Appendix A.1.1 of \cite{cubuk2020randaugment}). 

This simple RandAugment policy requires no search and yet achieves test set accuracies, on the aforementioned data sets and model, that are on-par or better than the more complex techniques previously discussed. To the best of our knowledge, only \AAA{} \cite{zhang2019adversarial} has significantly outperformed RandAugment. 

In Tab. \ref{tab:literature} we compare the test set accuracies (as reported by each  original publications) on CIFAR-10 and CIFAR-100 using Wide-ResNet-28-10 when using a baseline technique (Cutout) as well as some of the most recent proposals (among which RandAugment and \AAA{}).

\section{The Adversarial Approach to Image Augmentation}
\label{sec:Adversarial}

The core idea of \AAA{} consists of finding those augmentations that lead to ``hard'' samples, as measured by their training loss. The hypothesis is that learning from hard examples should force the model to learn more robust features \cite{zhang2019adversarial}. The  optimization objective becomes a min-max problem:
\begin{equation}
    \min_{w} \max_{\tau \in \mathcal{S}} \mathbb{E}_{x,y \sim P(\mathcal{D})} [\mathcal{L}(f_{w}(\tau(x),y))],
\label{eq:advobj}
\end{equation}
where, $\mathcal{D}$ is the training set, $x$ is a sample with label $y$, $w$ are the model parameters, $f_w(\cdot)$ is the model output, $\mathcal{L}(f_{w}(\cdot))$ is the loss function, and $\mathcal{S}$ is the set of all available augmentations (as in AutoAugment). In \cite{zhang2019adversarial} the inner maximization (i.e., find the most adversarial policies) is solved by introducing a controller augmentation policy network that, given the losses produced by the main model, is trained to output probabilities $p_1, \ldots, p_{|S|}$ for each $\tau_1, \ldots, \tau_{|S|}$, such that the maximization is solved in expectation (combining Eqns. (3) and (7) in \cite{zhang2019adversarial}):
\begin{equation}
        \min_{w } \max_{\theta} \mathbb{E}_{x,y\sim P(\mathcal{D})} \mathbb{E}_{\tau \sim P(\cdot, \theta)}[\mathcal{L}(f_{w}(\tau(x),y))],
\label{eq:advobjstoch}
\end{equation}
where $P(\cdot, \theta)$ is determined by the output of the controller policy network with parameters $\theta$.

\begin{table}[h]
  \caption{
Test set accuracy reported in recent works on automated image augmentation policy search for Wide-ResNet-28-10. Importantly, only \AAA{} consistently outperform RandAugment. 
}\label{tab:literature}
  \centering
  \begin{tabular}{lcc}
    \toprule
    & \multicolumn{2}{c}{Data set}                   \\
    \cmidrule(r){2-3}
    Work     & CIFAR10     & CIFAR100 \\
    \midrule
    Baseline (Cutout \cite{devries2017improved}) & 96.9  & 81.6     \\
    AutoAugment \cite{cubuk2019autoaugment} & +0.4  & +1.3   \\
    Fast AutoAugment \cite{lim2019fast} & +0.4 & +1.3      \\
    Faster AutoAugment \cite{hataya2020faster}    & +0.5       & +1.1  \\
    PBA \cite{ho2019population}    & +0.4      & +1.7  \\
    RandAugment \cite{cubuk2020randaugment}     & +0.4       & +1.7  \\
    \AAA{} \cite{zhang2019adversarial}    & +1.2       & +2.9  \\
    \bottomrule
  \end{tabular}
\end{table}

The approach in \cite{zhang2019adversarial}, summarized by Eq.~(\ref{eq:advobjstoch}), significantly outperforms the other works on automated image augmentation (Tab. \ref{tab:literature}). Yet, the training of the controller augmentation policy network introduces a large amount of stochasticity due to: (i) the training dynamics inside the min-max problem and (ii) the REINFORCE algorithm \cite{williams1992simple} necessary to cope with the non-differentiable (discrete) augmentation operations in the objective. 
Hence, it is logical to assume that the policy network will act as a random sampling algorithm, at least until it has learnt how to correctly predict which policies are the hardest for the given model. To complicate things, the model being trained is constantly evolving making the optimal policy a moving target. Therefore, it is unclear that the policies selected by the controller network are strictly adversarial. 

In Section~\ref{sec:Experiments}, we focus on determining whether being truly adversarial actually helps the model learning more robust features, or if there are more complex dynamics that should be considered.  

\section{An Adversarial Approach}
\label{sec:Experiments}

All our experiments are obtained following exactly the same training protocol, which is described below and very similar to \cite{zhang2019adversarial}. For computational reasons we discretize the magnitudes to $5$ values rather than $10$, and reduced the search space of the operations to $15$, as in \cite{ho2019population}. We use an initial batch size of 128, we adopt a multiplicity of $M=1$ and $M=2$, that is augmenting each batch using one or two operations (the latter approach effectively doubling the batch size as in \cite{zhang2019adversarial}). 
We train for 200 epochs and report the best test-accuracy over all training epochs, averaged across five runs. The learning rate is set to 0.1 with decay of 0.2 after the 50th, 100th, and 140th epochs, we use a warm-up of 10 epochs in which no augmentation is used, Nesterov optimizer, batch-norm momentum set to 0.1, and weight decay set to 5$e$-4. We trained the model using a cluster equipped with Nvidia GPUs model Tesla V100-SXM2-32GB with Cuda 10.1. The code was developed using Pytorch 1.8.1. Given the slightly different setup, results in Tab.~\ref{tab:experiments} are not meant for a direct comparison with \cite{zhang2019adversarial}. They can instead be compared with each other since we have followed the same training protocol.

\subsection{Removing the Stochasticity of the Controller Augmentation Policy Network Training}

We note that the optimal solution to Eq.~(\ref{eq:advobjstoch}) is determined by the optimal solution to Eq.~(\ref{eq:advobj}).

\begin{corollary}1
Let $\tau^* = \tau_j$ be the optimal solution to the inner maximization in Eq.~(\ref{eq:advobj}) for some $j \leq |\mathcal{S}|$ (for simplicity we assume only one optimum). Then any optimal solution $\theta^*$ to the inner maximization in Eq.~(\ref{eq:advobjstoch}) has to fulfill $p_j(\cdot, \theta^*) = 1$.
\end{corollary}

Corollary 1 can be proven by using Fubini-Tonelli to interchange the expectations in Eq.~(\ref{eq:advobjstoch}). The extension to multiple optima would simply share the probability mass among those. We run experiments on CIFAR10 and CIFAR100 with a Wide-ResNet-28-10 in which we directly solve for $\tau^*$ in Eq.~(\ref{eq:advobj}) during training to remove the stochasticity from the learning dynamics of the controller augmentation policy network. 

\textit{TrueAdv} refers to an experiment where in order to find the \textit{true} $\tau^*$, at the end of each epoch the empirical loss is evaluated on the entire training set and all the polices. Hence for all training points $N$ and for each policy $\tau \in \mathcal{S}$ we evaluate: $\frac{1}{N}\sum_{i=1}^N \mathcal{L}(f_{w}(\tau(x_i),y_i) \approx \mathbb{E}_{x,y \sim P(\mathcal{D})} [\mathcal{L}(f_{w}(\tau(x),y)]$. This algorithm is clearly not scalable but it is used to gain more insights on the matter.

\begin{figure}[h]
  \centering
  \subfigure[Operations]{\includegraphics[width=0.44\linewidth]{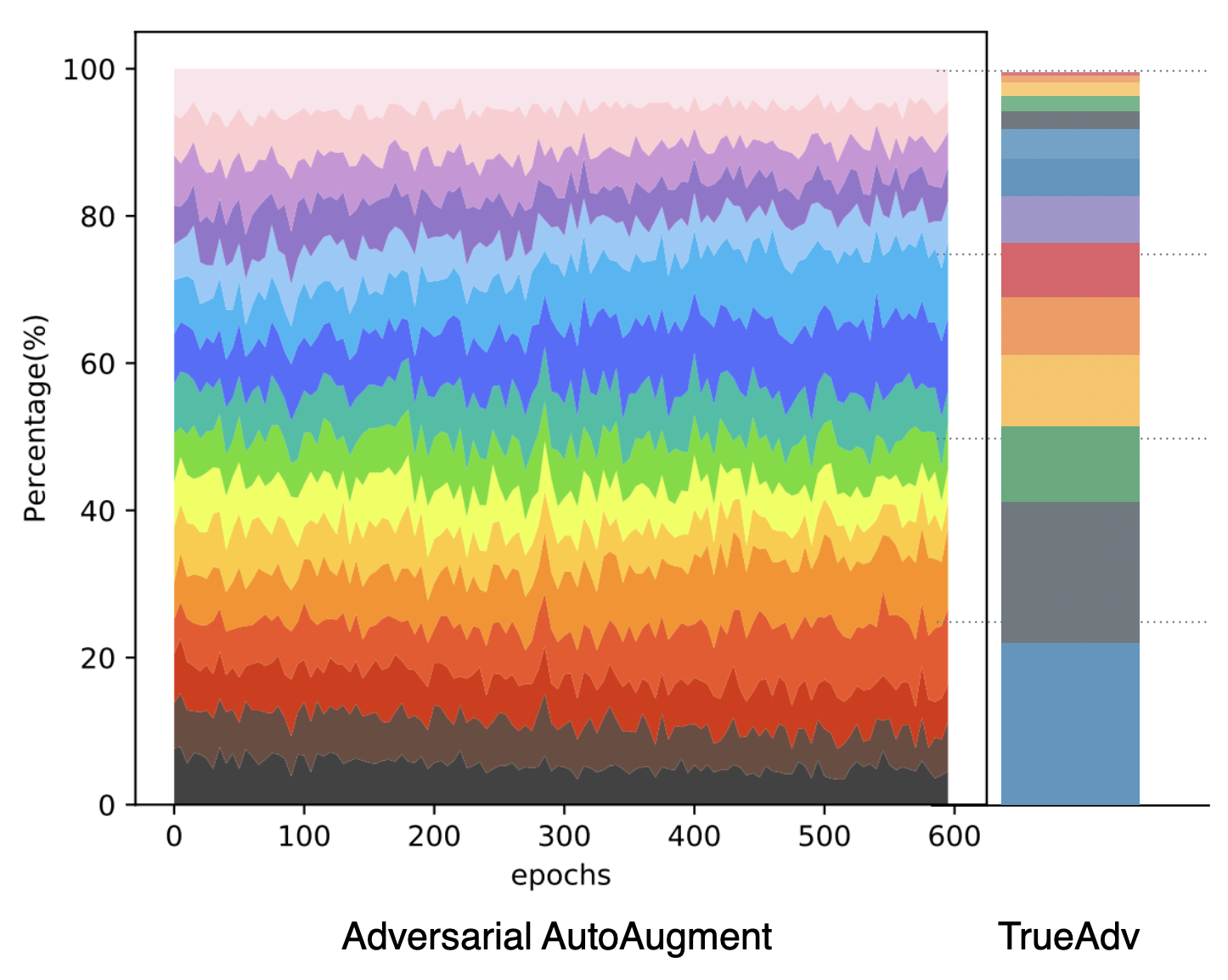}}
  \subfigure[Magnitudes]{\includegraphics[width=0.54\linewidth]{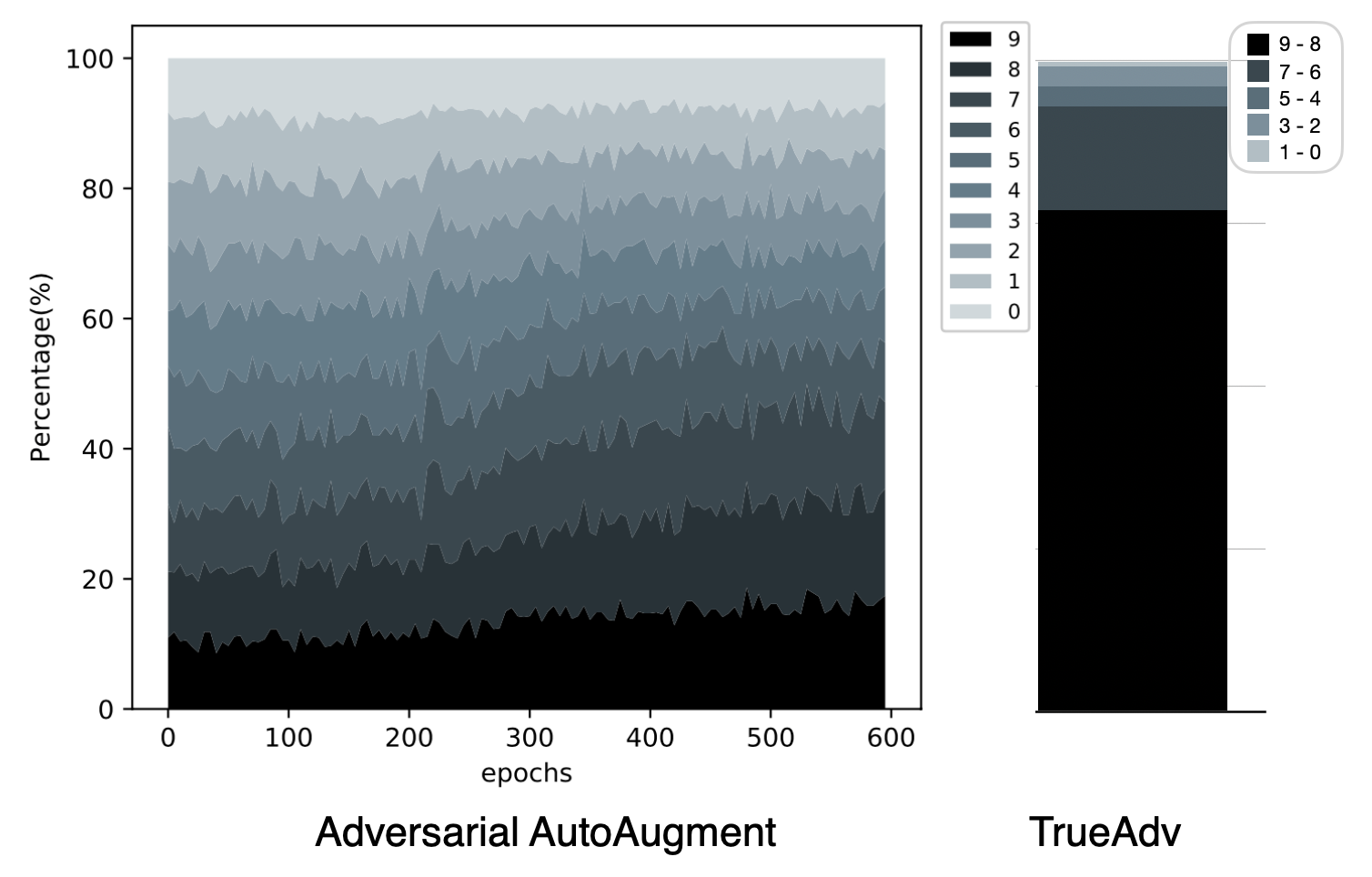}}
  \caption{Comparison between \AAA{} (as reported in \cite{zhang2019adversarial}) and \textit{TrueAdv}. We show the percentage of times each operation or magnitude were used during training. For \textit{TrueAdv} we report the average over all epochs, and magnitudes are binned by increasing strength (i.e., bin 8-9 is the highest magnitude). Notice how the operations (a) in \AAA{}  maintains an approximately uniform sampling across epochs, while \textit{TrueAdv} relies more often on fewer operations. Also, notice in (b) how the top two highest magnitudes (8-9) are selected less than 30\% of the time by \AAA{} but 77\% by \textit{TrueAdv}. If we analyze only the first 200 epochs (the stopping point for \textit{TrueAdv}) this difference would be amplified. 
  \vspace{-2mm}
  }
  \label{fig:augs_comparison}
\end{figure}

\begin{figure}[h]
  \centering
  \includegraphics[width=0.45\linewidth]{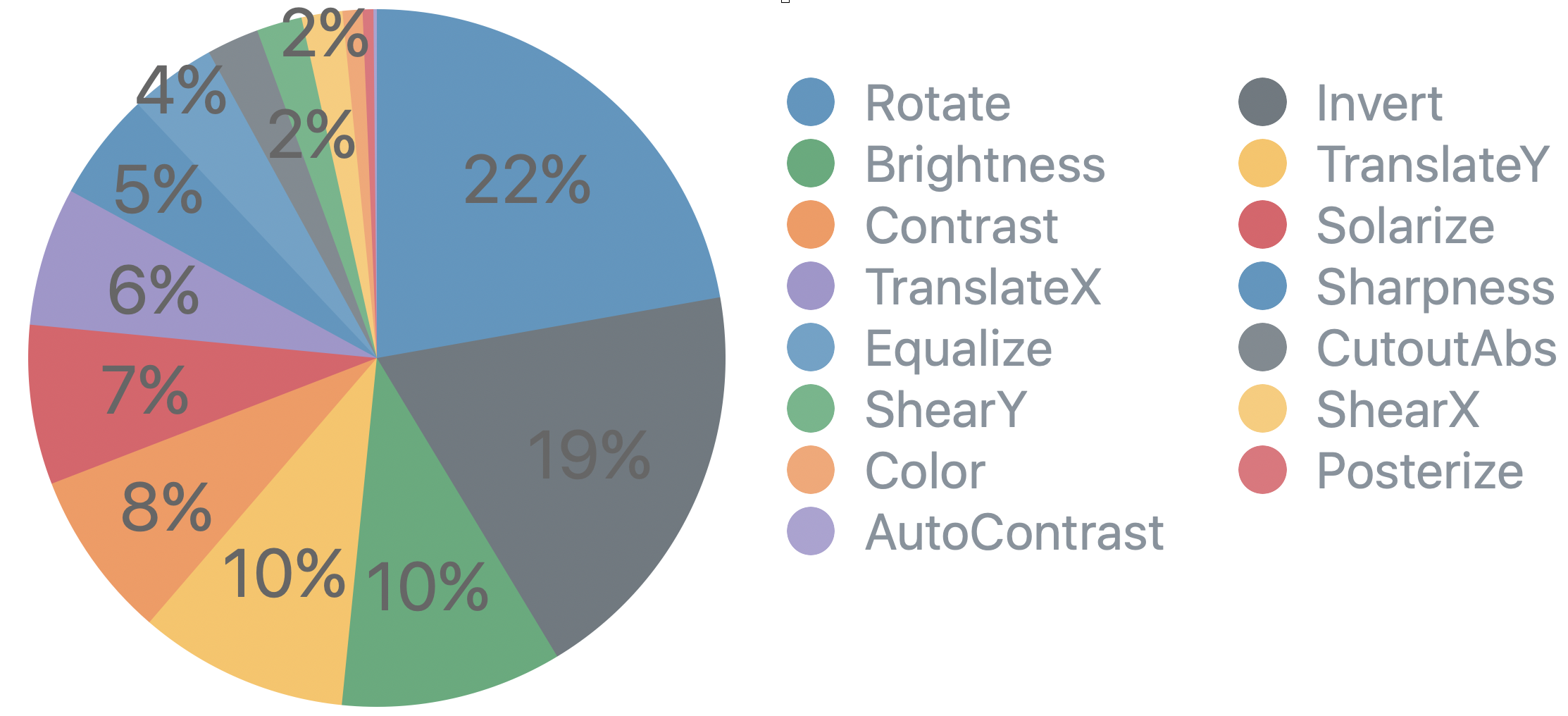}
  \hspace{1cm}
  \includegraphics[width=0.30\linewidth]{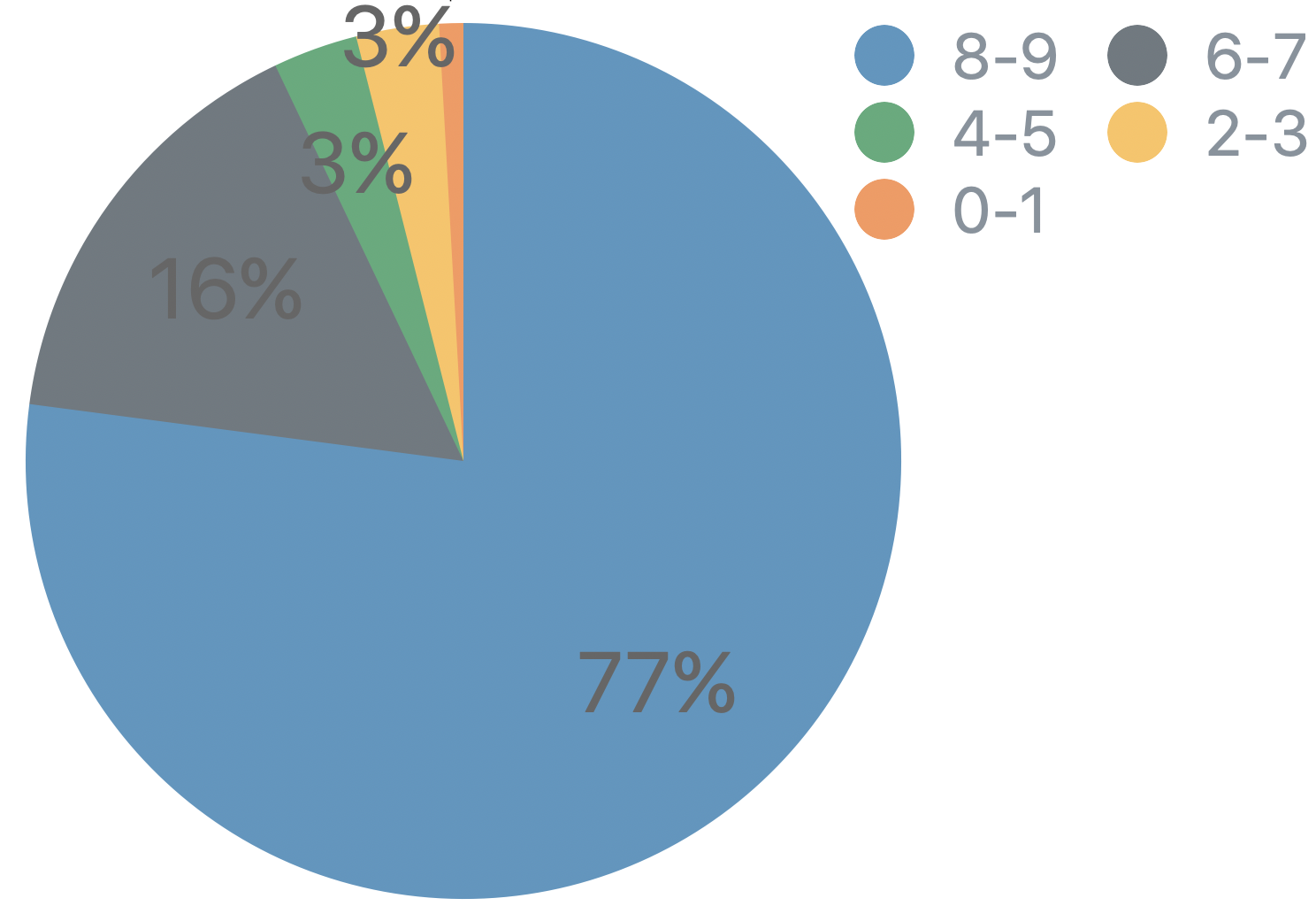}
  \caption{Percentage of times a given operation and magnitude were used during training (\textit{TrueAdv} strategy). Magnitudes have been binned into increasing strength (i.e., bin 8-9 is the highest magnitude, while 0-1 is the lowest magnitude).}
  \label{fig:augs}
\end{figure}

As shown in Tab.\ref{tab:experiments}, \textit{TrueAdv} is actually harmful to generalization, as it lowers test set accuracy for all data sets and multiplicities analyzed. Choosing random augmentations is not only computationally more efficient, but also better in terms of generalization. This leaves the question ``why do the learned policies in \cite{zhang2019adversarial} not exhibit the same behavior, despite aiming to solve an objective whose exact solution degrades generalization performance''? We hypothesize that this is due to the stochasticity of the policy network training. 
Fig.~\ref{fig:augs_comparison} shows the operations and magnitudes used by \AAA{} and \textit{TrueAdv} (details about the \textit{TrueAdv} breakdown can be found in Fig. \ref{fig:augs}). The differences in the choice of the policies are evident: (i) \textit{TrueAdv} relies only on 3 operations (Rotation, Inversion, Brightness) for more than 50\% of the time, while for \AAA{} the probability of sampling a specific operation remains fairly uniform (some milder differences can be appreciated after epoch 300); (ii)  \textit{TrueAdv} uses the highest magnitudes (8-9) 77\% of the time compared to the 30\% of \AAA{}.

\begin{table}
  \caption{
Test set accuracy gains over the baseline with M=1 (our implementation of Cutout \cite{devries2017improved}) with Wide-ResNet-28-10 trained for $200$ epochs. Mean of $5$ runs, standard deviations in brackets). 
}\label{tab:experiments}
  \centering
  {\fontsize{8}{7.2}\selectfont
  \begin{tabularx}{\textwidth}{lc|ccccccc}
    \toprule
    Accuracy \%               &     & Baseline & Random             & TrueAdv    & 1-Adv-0Ep & 1-Adv-100Ep & Smooth & Cyclic\\
    \midrule
    \multirow{2}{*}{CIFAR10} & M=1 &  96.8 (0.1) &\textbf{+0.5 (0.0)}& -1.1 (0.2)      & -          & -                           & +0.3 (0.0) & \textbf{+0.5 (0.1)}\\
                             & M=2 & - & +0.8 (0.1)         & -          & +0.8 (0.1) & +0.7 (0.0)   & +0.8 (0.1) &\textbf{+0.9 (0.1)}\\
    \multirow{2}{*}{CIFAR100}& M=1 & 81.1 (0.2) &\textbf{+1.3 (0.1)}& -0.8 (0.3) & -          & -            & +0.7 (0.1)  & +1.1 (0.2)\\
                             & M=2 & -  &\textbf{+1.8 (0.3)}& -          & +0.9 (0.1) & +0.9 (0.2)   & +0.6 (0.3) &+1.2 (0.4)\\
    \bottomrule
  \end{tabularx}
  }
\end{table}

These observations suggest that \AAA{} is not maximizing the inner objective of Eq.~\ref{eq:advobjstoch}, rather it samples reasonably uniformly. A form of curriculum could also be playing a role, however, the probabilities of sampling hard policies remain fairly low compared to \textit{TrueAdv}.

Leveraging these observations, we experiment with 4 different curricula. To make the approaches computationally tractable (unlike \textit{TrueAdv}), we evaluate the losses at the end of each epoch on a randomly sampled batch of size 128 and on a random subset of 500 policies. \textit{1-Adv-0Ep} uses multiplicity $M=2$, one hard augmentation (i.e. $\tau^*$, the augmentation with the highest average loss), and one augmentation that is sampled from the $250$ with the lowest average loss (i.e. easiest). \textit{1-Adv-100Ep} begins by using the easiest augmentations until epoch $100$, and then switches to  \textit{1-Adv-0Ep}. \textit{Smooth} gradually increases the adversarial approach starting with the easiest augmentations, the first hard augmentation (i.e. $\tau^*$) is introduced in epoch $75$ (when $M=2$) or epoch $125$ (when $M=1$). For $M=2$, the augmentation with the second highest loss is added at epoch $150$. 

All three curricula improve over the baseline, as shown in Tab.\ref{tab:experiments}, but cannot exceed the simple random policy. We notice how \textit{TrueAdv} suffers from over-fitting to the training data, suggesting that truly adversarial augmentations are creating a distribution shift away from \emph{real} data. Therefore, we test a curriculum that reverts back to the easiest augmentations at the end of training, to match the \emph{actual} expected distribution.
\textit{Cyclic} starts with the easiest augmentations,  for $M=2$, the first hard augmentation is introduced at epoch $75$, the second at epoch $100$. At epoch  $125$ it reverts back to only one hard augmentation, and finally from epoch $150$ on-wards it uses only the easiest set. For $M=1$ it introduces the hard augmentation at epoch $75$ before going back to the easy random augmentations at epoch $150$. Interestingly, the \textit{Cyclic} schedule achieves strong improvements over the baseline augmentation (Tab.\ref{tab:experiments}), even slightly surpassing random policies on CIFAR10. Random is still better, however, on CIFAR100, especially in the case of $M=2$.

\section{Discussion}
\label{sec:Discussion}
In this work we showed that when choosing training augmentations solving exactly the adversarial objective introduced in \cite{zhang2019adversarial} deteriorates the test set accuracy, even when compared to the baseline \cite{devries2017improved}. We believe that the results achieved in \cite{zhang2019adversarial} must come from its stochasticity and a mild form of curriculum that emerges from the network policy controller. Our attempts to find an explicit curriculum, however, led to improve the test set accuracy compared to the baseline but not compared to a random augmentation policy. The idea of reverting towards milder augmentations as the model completes the training has proven to be interesting, and in future works we will continue exploring how this strategy can be refined to finally produce results that are better than using random policies, while automatizing as much as possible the hyper parameters of the curriculum.

\newpage
\bibliographystyle{plain}
\bibliography{ms}

\end{document}